\documentclass[sigconf]{acmart}
\AtBeginDocument{%
  }

\copyrightyear{2026}
\acmYear{2026}
\setcopyright{cc}
\setcctype{by}
\acmConference[RecSys '26]{20th ACM Conference on Recommender Systems}{September 27-October 02, 2026}{Minneapolis, MN, USA}
\acmBooktitle{20th ACM Conference on Recommender Systems (RecSys '26), September 27-October 02, 2026, Minneapolis, MN, USA}
\acmDOI{10.1145/3773078.3831929}
\acmISBN{979-8-4007-2284-4/2026/09}

\usepackage{multirow}
\graphicspath{{./}{Figures/}}
\begin{document}

\title[LLM Reasoning for Subjective Tasks]{LLM Reasoning for Subjective Tasks: Failure Modes, Mitigation, and Dynamic Reasoning Routing}

\author{Juncheng Dong}
\authornote{This work was conducted during JD's internship at Netflix.}
\affiliation{%
  \institution{Duke University}
  \city{Durham}
  \country{USA}}
\email{juncheng.dong@duke.edu}

\author{Ding Tong}
\affiliation{%
  \institution{Netflix}
  \city{Los Gatos}
  \country{USA}}
\email{dingt@netflix.com}

\author{Ishan Gupta}
\authornote{Work done at Netflix.}
\affiliation{%
  \city{Los Gatos}
  \country{USA}}
\email{345ishaan@gmail.com}

\author{Yuyan Wang}
\affiliation{%
  \institution{Netflix}
  \city{Los Gatos}
  \country{USA}}
\email{yuyanw@netflix.com}

\renewcommand{\shortauthors}{Dong et al.}

\begin{abstract}
Recommendation systems thrive on personalization, where ``correctness'' is rarely a binary truth but a matter of subjective human preference. As Large Language Models (LLMs) are deployed as autonomous verifiers of safety and quality guidelines, they face a distinctive challenge: context-aware preference alignment. Recent gains in Reinforcement Learning with Verifiable Rewards (RLVR) are indexed mostly on objective, mathematical tasks. Through a large-scale study spanning both proprietary and open-source models on four real-world verification tasks from a production recommender platform, we ask whether explicit reasoning generalizes to subjective, human-centric industry rubrics. We expose a fundamental vulnerability: rigid, math-centric reasoning traces actively degrade verification, and applying standard RLVR triggers a phenomenon we term \textit{reasoning collapse}, in which the policy abandons deliberation in favor of rapid heuristic guessing. We introduce a conditional length-penalized post-training algorithm that intertwines verification accuracy with bounded reasoning length, halting collapse and recovering performance. Finally, we show that a reasoning trace's efficacy is tightly coupled with its socio-linguistic framing: across 1{,}500 synthesized personas, verification accuracy swings by nearly $0.38$ macro-F1 depending solely on the adopted reasoning persona---evidence that much subjective-verification error is really reasoning-style mismatch. This observation motivates a mid-training architecture that routes reasoning through contextually aligned personas. This work offers both a scalable algorithmic patch and a long-term architectural blueprint for aligning reasoning models with real-world subjective constraints.
\end{abstract}


\maketitle

\section{Introduction}

Large language model (LLM) reasoning has fundamentally expanded model capabilities, enabling breakthroughs in complex, objective challenges such as mathematical proofs and code generation \citep{Lightman2023, Uesato2022, Cobbe2021}. By generating intermediate thinking tokens, e.g., Chain-of-Thought (CoT), LLMs increase the likelihood of reaching the correct final answer~\citep{Wei2022}. Recently, a paradigm shift has shown that models can self-learn these reasoning trajectories through Reinforcement Learning with Verifiable Rewards (RLVR): leveraging algorithms like Group Relative Policy Optimization (GRPO), LLMs internalize reasoning purely through trial, error, and outcome supervision \citep{DeepSeekR1_2025, OpenAIo1_2024, Shao2024}. Crucially, this success is anchored in objectivity, relying on deterministic verifiers such as compilers for robust reward signals.

\begin{figure*}[t]
    \centering
    \includegraphics[width=0.86\linewidth]{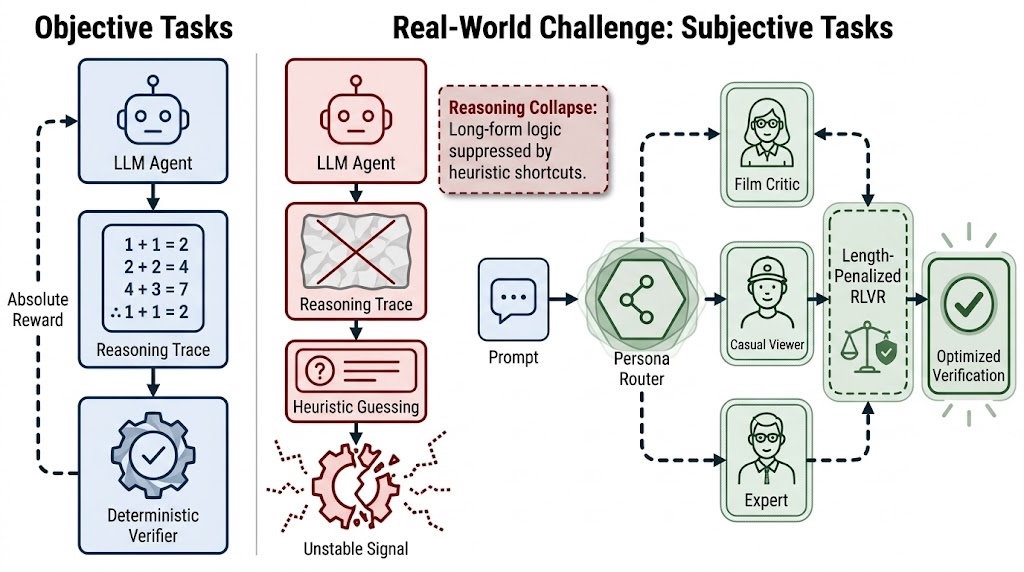}
    \caption{(Left) RLVR enhances reasoning when pre-training contains relevant reasoning traces and verifiers are reliable. (Middle) RecSys instead faces subjective tasks \emph{without} correct traces or reliable verifiers, where straightforwardly applying RLVR induces a critical \emph{reasoning collapse}. (Right) We propose a length-penalized RLVR algorithm to address reasoning collapse, plus persona-driven reasoning, where the LLM first learns diverse reasoning biases from different personas and then in-context selects the appropriate bias for each subjective task.}
    \Description{Three-panel overview: (left) standard RLVR with reliable reasoning traces and verifiers; (middle) subjective RecSys tasks lack reliable traces and verifiers, so naive RLVR causes reasoning collapse; (right) our length-penalized RLVR plus persona-driven reasoning.}
    \label{fig:schema}
\end{figure*}

In contrast, modern recommender systems (RecSys) operate in domains that are inherently subjective and human-centric \citep{Hou2023, Bao2023, Kang2023}. As the industry rapidly adopts ``LLM-as-a-Judge'' frameworks for autonomous content moderation and personalized query verification \citep{Zheng2023, Dubois2024, WangAlign2023}, these models must evaluate abstract guidelines like cultural sensitivity and conversational tone, where reliable verifiers are difficult to define. A critical open question remains: \textit{can the explicit reasoning techniques forged in objective domains generalize to subjective industry tasks?} An overview of our setting and contributions is shown in Figure~\ref{fig:schema}.

This work investigates this generalization gap through a large-scale empirical analysis using real-world datasets from Netflix. Surprisingly, our evaluations reveal that forcing LLMs to utilize explicit reasoning on subjective verification tasks is frequently unhelpful and often actively degrades performance, across both proprietary and open-source models. When applying standard RLVR (e.g., GRPO) to rectify this, the models fail to improve. In fact, if a model already exhibits baseline reasoning proficiency, applying subjective RLVR often diminishes its accuracy---a failure mode echoing the known vulnerabilities of reward overoptimization and misalignment in RLHF \citep{Casper2023, Gao2023}. While RLVR succeeds in mathematics because the pre-training corpus is saturated with step-by-step proofs, subjective domains lack a canonical ``correct'' reasoning trajectory, causing the reinforcement signal to falter.

Most importantly, we uncover a counterintuitive training dynamic that we term \textit{reasoning collapse}. Because subjective domains lack a canonical ``correct'' trajectory, the LLM learns to optimize expected reward faster via short-form heuristic guessing than long-form deliberation; consequently, extended reasoning is rapidly suppressed by the policy. To overcome this, we propose a length-penalized post-training algorithm that applies reward shaping exclusively to correct answers bounded by a dynamic length limit, stabilizing the reasoning trajectory and recovering task performance. While this resolves the mechanical collapse, it exposes a deeper limitation: a ``reasoning mismatch'' where LLMs erroneously apply analytical mathematical logic to human-centric problems. We therefore further explore \emph{persona-driven reasoning}: prior work shows LLMs can simulate distinct generative agents \citep{Park2023, Shanahan2023, Salewski2023}, and our preliminary synthesis demonstrates that routing reasoning through a contextually aligned persona shows significant promise for subjective verification, motivating a framework where RLVR teaches an LLM \textit{which} reasoning bias to apply.

\paragraph{Contributions.} (i) A large-scale empirical study showing that explicit and intrinsic reasoning frequently \emph{degrade} subjective verification, across proprietary and open-source models (Section~\ref{sec:empirical_eval}); (ii) the identification and characterization of \emph{reasoning collapse} under standard RLVR on subjective tasks (Section~\ref{sec:reasoning_collapse}); (iii) a conditional length-penalized reward that halts collapse and recovers performance (Section~\ref{sec:algorithm}); and (iv) preliminary evidence that persona-conditioned reasoning drives large performance variance, motivating a persona-routing mid-training blueprint (Section~\ref{sec:exploratory}).

\section{Reasoning on Subjective Tasks}
\label{sec:empirical_eval}

We construct evaluations utilizing \emph{real-world datasets} from Netflix, where human annotators give subjective opinions. Unlike objective benchmarks grounded in mathematical truths, these tasks are abstract, rubric-driven, human-centric verifications critical for relevance, safety, and user satisfaction at scale.

\subsection{Tasks and Experimental Setup}
We evaluate four binary (``Yes''/``No'') verification tasks defined by internal safety and quality guidelines:
\begin{itemize}
    \item \textbf{(I) Conversational Query Sensitivity}
    \item \textbf{(II) Generated Text Response Sensitivity}
    \item \textbf{(III) Text: Response Quality}
    \item \textbf{(IV) ID: Response Quality}
\end{itemize}
Full rubric details are in Appendix~\ref{app:tasks}. For each task we collect queries and human ground-truth verifications. To isolate reasoning, we compare three prompt formulations:
\begin{itemize}
    \item a \emph{direct JSON verification} baseline without reasoning;
    \item \emph{implicit reasoning} (an explanation before the answer); and
    \item \emph{explicit reasoning} (structured \texttt{<reason>} and \texttt{<answer>} tags).
\end{itemize}
The proprietary reasoning model is additionally run at ``low'' and ``high'' effort. Unless noted, all numbers are macro-F1 averaged over 5 independent runs; the full evaluation and training protocol is given in Appendix~\ref{app:setup}.

\subsection{Proprietary Model Degradation}
We first benchmark two state-of-the-art proprietary models: a frontier general-purpose LLM (Frontier-LLM) and a frontier reasoning model (Frontier-Reasoner). As detailed in Table~\ref{tab:proprietary_performance}, we observe a \emph{counterintuitive trend}: explicitly prompting for reasoning, or using models built for deep reasoning, generally degrades verification relative to the zero-shot baseline.

\begin{table*}[t]
\centering
\caption{Performance (macro-F1, averaged over 5 runs) of prompting-based verifiers on proprietary models. Frontier-LLM denotes a state-of-the-art proprietary general-purpose model; Frontier-Reasoner denotes a state-of-the-art proprietary reasoning model. Enforcing explicit reasoning or high-effort intrinsic reasoning degrades performance across most subjective tasks.}
\label{tab:proprietary_performance}
\renewcommand{\arraystretch}{1.05}
\small
\begin{tabular}{@{}lccccc@{}}
\toprule
\multirow{2}{*}{\textbf{Task Dataset}} & \multicolumn{3}{c}{\textbf{Frontier-LLM}} & \multicolumn{2}{c}{\textbf{Frontier-Reasoner}} \\ \cmidrule(lr){2-4} \cmidrule(lr){5-6}
 & \textbf{w/o Rea} & \textbf{Imp Rea} & \textbf{Exp Rea} & \textbf{low} & \textbf{high} \\ \midrule
Query Sensitivity      & \textbf{0.901} & 0.893 & 0.887 & 0.895 & 0.874 \\
Response Sensitivity   & \textbf{0.622} & 0.605 & 0.619 & 0.564 & 0.580 \\
\;\;\;\;Self-Consistency ($K{=}8$)      & \textbf{0.690} & 0.593 & 0.574 & 0.615 & 0.623 \\
Text: Response Quality    & \textbf{0.892} & 0.884 & 0.876 & 0.882 & 0.863 \\
ID: Response Quality                    & 0.513 & 0.517 & \textbf{0.520} & 0.503 & 0.502 \\ \bottomrule
\end{tabular}
\end{table*}

For Task~I, the general-purpose model without reasoning achieves macro-F1 $0.901$; explicit reasoning reduces this to $0.887$, and Frontier-Reasoner (high effort) is worse at $0.874$. Task~II shows the same: $0.622$ without reasoning, $0.619$ with explicit reasoning, and $0.564$ for the reasoning model (low effort). Task~III follows the same pattern; reasoning helps the general-purpose model only on Task~IV, and even there it \emph{degrades} the dedicated reasoning model---an inconsistency that casts further doubt on transferring objective-task reasoning to subjective verification. Thus, for proprietary models heavily aligned for objective problem-solving, intermediate reasoning appears to introduce noise and hallucinations that disrupt subjective evaluation. Inference-time scaling does not rescue it: on Task~II, even Majority Vote ($K{=}8$) falls from $0.690$ (no reasoning) to $0.574$ (explicit)---a larger gap---likely because voting reinforces unhelpful reasoning.

\subsection{The Open-Source Reasoning Paradox}
To determine whether this is an artifact of proprietary alignment, we evaluate Mistral-7B-Instruct (weaker mathematical reasoning) and Qwen2.5-7B-Instruct (strong alignment and mathematical reasoning). Table~\ref{tab:opensource_performance} reports baselines before reinforcement learning: a clear dichotomy emerges. Mistral-7B benefits from reasoning on every task, whereas the effect on Qwen2.5 is inconsistent and task-dependent---reasoning helps on Query Sensitivity but degrades Response Sensitivity.

\begin{table*}[t]
\centering
\caption{Macro-F1 (averaged over 5 runs) of open-source models across subjective tasks. Qwen2.5 shows mixed results when reasoning is enforced; Mistral-7B universally benefits.}
\label{tab:opensource_performance}
\renewcommand{\arraystretch}{1.05}
\small
\begin{tabular}{@{}lcccc@{}}
\toprule
\multirow{2}{*}{\textbf{Task Dataset}} & \multicolumn{2}{c}{\textbf{Mistral-7B-Instruct}} & \multicolumn{2}{c}{\textbf{Qwen2.5-7B-Instruct}} \\ \cmidrule(lr){2-3} \cmidrule(lr){4-5}
 & \textbf{w/o Reason} & \textbf{With Reason} & \textbf{w/o Reason} & \textbf{With Reason} \\ \midrule
Query Sensitivity & 0.707 & \textbf{0.724} & 0.748 & \textbf{0.805} \\
Response Sensitivity & 0.555 & \textbf{0.646} & \textbf{0.561} & 0.520 \\
Text: Response Quality& 0.402 & \textbf{0.465} & 0.458 & \textbf{0.512} \\
ID: Response Quality & 0.505 & \textbf{0.518} & \textbf{0.544} & 0.539 \\ \bottomrule
\end{tabular}
\end{table*}

This pattern highlights a \emph{reasoning mismatch}. Models already strong at math-centric reasoning (Qwen, and both proprietary models) transfer that reasoning to human-centric rubrics only unreliably---helping on some tasks yet degrading others---whereas the weaker-math Mistral consistently turns the context window into flexible scratchpad space for a steady gain. Standard reasoning learned for math and code thus does not reliably transfer to subjective industry tasks.

\section{Post-Training and Reasoning Collapse}
\label{sec:reasoning_collapse}

Given the baseline degradation, a natural countermeasure is to post-train with RLVR, which in objective domains incentivizes models to refine their own reasoning. We hypothesize GRPO might force the LLM to learn a more appropriate, task-specific reasoning pattern.

\subsection{Inefficacy of RLVR Variants}
We post-train Qwen2.5-7B-Instruct and Mistral-7B-Instruct, expanding beyond standard GRPO~\citep{DeepSeekR1_2025} to Dr.~GRPO, which removes length and standard-deviation normalization to address optimization bias~\citep{liu2025understandingr1zeroliketrainingcritical}, and GSPO, which operates at the sequence rather than token level to reduce variance~\citep{zheng2025groupsequencepolicyoptimization}. Rewarding strictly on the final verification label (training details in Appendix~\ref{app:setup}; results averaged over 5 runs), all three algorithms yield counterintuitive outcomes (Table~\ref{tab:rlvr_performance}) that expose a key limitation of monolithic reasoning traces.

\begin{table*}[t]
\centering
\caption{Post-training reasoning performance (macro-F1, averaged over 5 runs). Advanced algorithms (Dr.~GRPO, GSPO) fail to rescue Qwen's collapse, while Mistral shows consistent gains.}
\label{tab:rlvr_performance}
\renewcommand{\arraystretch}{1.05}
\small
\begin{tabular}{@{}llcccc@{}}
\toprule
\textbf{Task Dataset} & \textbf{Model} & \textbf{Base (Reason)} & \textbf{GRPO} & \textbf{Dr.~GRPO} & \textbf{GSPO} \\ \midrule
\multirow{2}{*}{Query Sensitivity} & Mistral & 0.724 & 0.765 & 0.781 & \textbf{0.802} \\
 & Qwen & \textbf{0.805} & 0.749 & 0.758 & 0.762 \\ \midrule
\multirow{2}{*}{Response Sensitivity} & Mistral & 0.646 & 0.672 & 0.688 & \textbf{0.705} \\
 & Qwen & \textbf{0.520} & 0.493 & 0.498 & 0.504 \\ \midrule
\multirow{2}{*}{Text: Response Quality} & Mistral & 0.465 & 0.490 & 0.505 & \textbf{0.520} \\
 & Qwen & 0.512 & 0.558 & 0.561 & \textbf{0.565} \\ \midrule
\multirow{2}{*}{ID: Response Quality} & Mistral & 0.518 & 0.535 & 0.542 & \textbf{0.550} \\
 & Qwen & \textbf{0.539} & 0.519 & 0.522 & 0.526 \\ \bottomrule
\end{tabular}
\end{table*}

\textbf{The Monolithic Reasoning Paradox.}
The divergence between Qwen and Mistral is telling. Qwen, pre-trained extensively on mathematics, holds a rigid, monolithic reasoning trace; under RLVR or its sequence-level variants its performance collapses. These algorithms fix mathematical optimization and length biases but cannot fix a socio-linguistic misalignment between a rigid math-proof structure and a human-centric rubric, so reward cannot cleanly credit intermediate thoughts. Mistral, with weaker mathematical reasoning, lacks this constraint and uses the context window as a flexible scratchpad, scaling better across all algorithms. Because the reward cannot cleanly attribute credit to individual intermediate thoughts, optimization drifts away from coherent deliberation and toward whatever shortcut maximizes the outcome signal. This supports our central hypothesis: monolithic, math-centric reasoning is not merely unhelpful but can be actively \emph{harmful} for subjective tasks. Succeeding on human-centric rubrics is therefore less a matter of better gradient clipping than of endowing models with dynamic, persona-aligned reasoning flexibility.

\subsection{Reasoning Collapse}
\begin{figure}[t]
\centering
\includegraphics[width=0.9\linewidth]{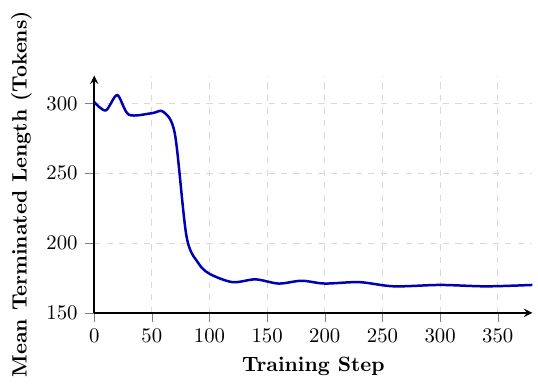}
\caption{Reasoning collapse during GRPO post-training: the mean terminated length of completions plummets after $\sim$70 steps as the model shortcuts to heuristic guessing.}
\Description{Line plot of mean terminated completion length versus training step; length stays near 300 tokens until about step 70, then drops sharply to a plateau near 170 tokens.}
    \label{fig:reasoning_collapse}
\end{figure}

The performance inversion produces a striking dynamic we term \textit{reasoning collapse}. In successful mathematical RLVR, the signature of emergent reasoning is a \emph{simultaneous increase} in response length and accuracy. We observe the opposite (Figure~\ref{fig:reasoning_collapse}): because the model \textit{without} reasoning improves faster during exploration than the model \textit{with} reasoning, the policy heavily penalizes long-form generation, and the mean terminated length plummets after $\sim$70 steps. Once the policy discovers that directly guessing the label earns the same---or higher---reward at a fraction of the token cost, it rapidly suppresses extended CoT and the reasoning trajectory collapses toward zero length. Standard outcome-based rewards thus appear insufficient to preserve explicit reasoning on subjective rubrics.

\section{Length-Penalized Post-Training}
\label{sec:algorithm}

To address the phenomenon of reasoning collapse, the reinforcement learning policy must be constrained to prevent the premature truncation of intermediate reasoning steps. Our primary objective is to force the LLMs to engage in extended reasoning during post-training without degrading the quality of the final verification output.

\subsection{Reward Shaping Formulations}
We model the post-training process within a standard reinforcement learning framework, where the base objective is to maximize the expected reward of the final generated token corresponding to the verification label. Let $r_{base} \in \{0, 1\}$ represent the objective binary reward, where $1$ indicates a correct verification and $0$ indicates an incorrect one. Let $L$ denote the token length of the generated reasoning response, and $L_{target}$ denote a hyperparameter representing the desired reasoning length capacity.

\textbf{Approach I: Unconstrained Length Rewarding.} Our initial attempt naively added a monotonically increasing length reward to the base objective. While this successfully prevented reasoning collapse, the policy exploited the reward function by generating repetitive, nonsensical tokens until reaching the absolute maximum context limit, collapsing the model's accuracy.

\textbf{Approach II: Target-Constrained Length Penalty.} To prevent infinite token generation, we introduced a bounded length penalty designed to smoothly guide the response length $L$ toward the target length $L_{target}$. The total reward $r_{total}$ was formulated as:
\begin{equation}
r_{total} = r_{base} + \lambda \min(0, L - L_{target})
\end{equation}
where $\lambda$ is a scaling constant. While this constraint successfully caused the response length to smoothly converge to $L_{target}$, empirical evaluations showed that the LLMs failed to utilize this extended generation window for accurate task-solving. The model generated coherent but functionally useless filler text before guessing the answer.

\textbf{Approach III: Conditional Length Rewarding (Proposed).} The core failure of the previous approaches was decoupling the length reward from the accuracy of the underlying reasoning. To ensure that the extended response length is meaningfully utilized to solve the task, the length reward must be strictly gated by the correctness of the final answer. We define our final conditional reward function as:
\begin{equation}
r_{total} =
\begin{cases}
r_{base} + \lambda L, & \text{if } r_{base}{=}1,\ L{<}L_{target} \\
r_{base}, & \text{otherwise}
\end{cases}
\end{equation}
Under this formulation, the model only receives the length bonus if it successfully arrives at the correct verification. Furthermore, capping the bonus at $L_{target}$ prevents catastrophic context exploitation.

\subsection{Recovering Performance}
Implementing our conditional length reward resolves reasoning collapse. As detailed in Table~\ref{tab:reward_shaping}, we track the progression of Qwen2.5-7B-Instruct across two distinct subjective rubrics: the Conversational Query Sensitivity task and the ID: Response Quality task. While Approach~I hallucinates infinitely and Approach~II generates unhelpful filler, our conditional algorithm translates the preserved reasoning capacity into consistent gains, stabilizing near the 1{,}000-token limit without degrading into gibberish. On the Query Sensitivity task---where reasoning is beneficial ($0.805$ vs.\ $0.748$ without)---naive GRPO collapses F1 to $0.749$, but our conditional reward lifts it to $0.851$ ($90.6\%$ accuracy), surpassing the reasoning baseline. Strikingly, the method rescues even the ID: Response Quality task, where reasoning is \emph{not} beneficial at baseline ($0.539$ vs.\ $0.544$): after GRPO collapses it to $0.519$, our reward recovers F1 to $0.572$---exceeding both the reasoning and no-reasoning baselines on this hardest rubric. By intertwining verification accuracy with reasoning length, the LLM learns to use explicit CoT for these subjective evaluations.

\begin{table}[t]
\centering
\caption{Reward-shaping formulations on Qwen2.5-7B-Instruct (macro-F1, averaged over 5 runs). The conditional length reward is the only formulation that preserves extended reasoning and recovers performance.}
\label{tab:reward_shaping}
\renewcommand{\arraystretch}{1.05}
\small
\begin{tabular}{@{}lcccc@{}}
\toprule
\multirow{2}{*}{\textbf{Configuration}} & \multicolumn{2}{c}{\textbf{Query Sens.}} & \multicolumn{2}{c}{\textbf{ID: Resp.\ Qual.}} \\ \cmidrule(lr){2-3} \cmidrule(lr){4-5}
 & \textbf{F1} & \textbf{Len.} & \textbf{F1} & \textbf{Len.} \\ \midrule
Base (no RLVR)             & 0.805 & $\sim$300 & 0.539 & $\sim$280 \\
GRPO (collapse)   & 0.749 & $\sim$170 & 0.519 & $\sim$150 \\ \midrule
Unconstrained (I)     & 0.210 & 8192 & 0.185 & 8192 \\
Target-constrained (II)      & 0.760 & $\sim$990 & 0.525 & $\sim$950 \\
\textbf{Conditional (III, ours)} & \textbf{0.851} & \textbf{$\sim$980} & \textbf{0.572} & \textbf{$\sim$960} \\ \bottomrule
\end{tabular}
\end{table}

\section{Dynamic Reasoning via Persona Mid-Training}
\label{sec:exploratory}

While our conditional reward mitigates the mechanical collapse, a deeper limitation remains: standard LLMs default to analytical, mathematics-based reasoning ill-suited for subjective tasks. We present preliminary findings pointing toward a dynamic, routing-based approach. Our key insight is that, since subjective tasks need different reasoning patterns than objective ones, we must first endow LLMs with diverse patterns---synthesizing and distilling them---and then use RLVR to in-context select the appropriate one. We hypothesize that different ``personas'' exhibit different reasoning patterns, as distinct groups of people express different thinking processes and opinions.

\subsection{Persona Impact on Verification}
\begin{figure}[t]
    \centering
    \includegraphics[width=\linewidth]{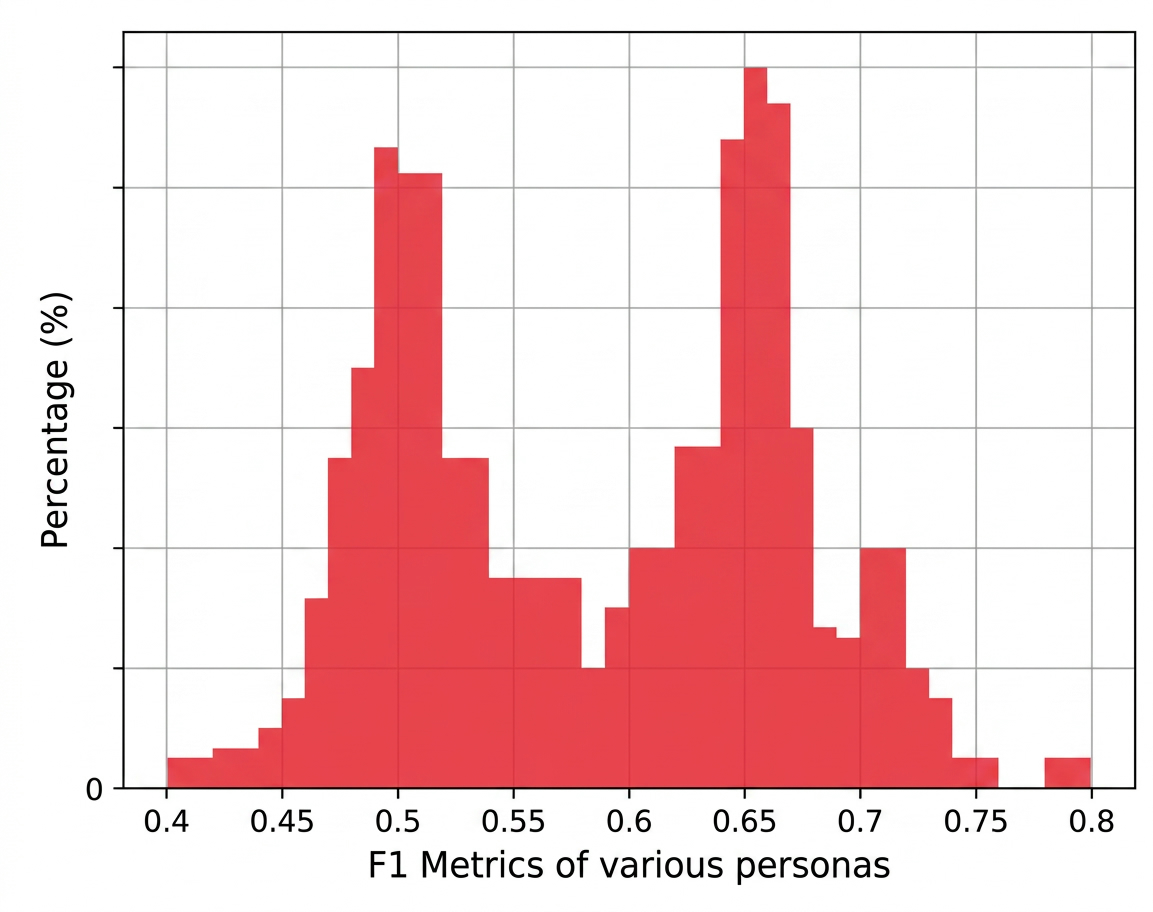}
    \caption{Reasoning patterns of different personas lead to drastically different verification performance (one sensitivity task; 1{,}500 personas).}
    \Description{Histogram of macro-F1 across 1,500 personas on a sensitivity task, a wide roughly bimodal spread from about 0.42 to 0.79.}
    \label{fig:persona}
\end{figure}

We synthesized varied reasoning trajectories using 1{,}500 distinct personas drawn from a public roleplay dataset~\citep{singh2026fspofewshotoptimizationsynthetic}, spanning a wide range of reasoning styles, perspectives, and communication norms. We prompted Frontier-LLM to impersonate each persona, generating reasoning traces and verification labels consistent with the assigned worldview (protocol in Appendix~\ref{app:persona}). Comparing these persona-conditioned labels against human ground truth, we observed large variance attributable to persona alignment, the only varied factor. As shown in Figure~\ref{fig:persona}, on one sensitivity task macro-F1 ranged from $0.416$ (worst persona) to $0.792$ (best)---a spread of nearly $0.38$ absolute F1 driven by which persona's reasoning the model adopted. The same wide spread holds on the harder ID: Response Quality task (Figure~\ref{fig:persona2}, Appendix~\ref{app:persona}). This suggests the ``worldview'' embedded in the CoT strongly shapes accuracy: a singular, math-aligned reasoning pattern is unlikely to serve all subjective verifications. Concretely, the best-performing persona nearly doubles the macro-F1 of the worst, indicating that a large share of what looks like ``verification error'' is really a \emph{reasoning-style mismatch} that an appropriately routed persona could recover.

\subsection{A Mid-Training Blueprint}
While deploying a fully realized persona-routing model is beyond this initial intervention, our synthesis suggests a promising path: not simply more supervised fine-tuning data, but a shift in how reinforcement learning is applied. We propose a three-step ``mid-training'' pipeline:
\begin{enumerate}
    \item \textbf{Synthesis} --- generate diverse Chain-of-Thought trajectories across thousands of distinct personas, building a broad repository of subjective reasoning patterns.
    \item \textbf{Supervised Fine-Tuning (SFT)} --- mid-train the base LLM to internalize this array of non-mathematical reasoning patterns.
    \item \textbf{Reinforcement Routing} --- whereas RLVR is historically used to teach an LLM \textit{how} to reason, we instead use GRPO to teach it \textit{which} reasoning bias to apply.
\end{enumerate}
The resulting LLM acts as a contextual multi-armed bandit: given a subjective query, the policy routes its reasoning through the persona node (e.g., ``empathetic moderator'' vs.\ ``strict policy enforcer'') that maximizes the verification reward, aligning verifiers with nuanced human rubrics.

This routing formulation has concrete advantages over a single monolithic trace. Because each persona encodes a distinct, human-interpretable stance, the router's choice is itself an explanation: an operator can see \emph{which} bias produced a verdict, audit it against policy, and retire personas that prove unreliable. The persona library can also be re-weighted as guidelines evolve, letting the verifier track shifting rubrics without full re-training. Key open questions---curating personas that exclude harmful or stereotyped stances, training the router with sufficient feedback, and evaluating worst-case behavior across personas---remain tractable engineering and governance challenges rather than fundamental barriers.

\section{Related Work}\label{sec:related-work}
Our contribution sits at the intersection of three lines of work, discussed in full in Appendix~\ref{app:relwork}. \emph{LLM reasoning and RL post-training} (CoT, GRPO and variants) has advanced rapidly \citep{Wei2022, DeepSeekR1_2025, OpenAIo1_2024, Shao2024} but is benchmarked almost entirely on objective, verifiable tasks \citep{Cobbe2021, Hendrycks2021}. \emph{Alignment under imperfect rewards} (RLHF, Constitutional AI, DPO, and RLVR) is known to suffer reward overoptimization on subjective feedback \citep{Ziegler2019, Bai2022, Rafailov2023, Casper2023, Gao2023}. \emph{LLM-as-a-Judge verification} is now standard in RecSys \citep{Hou2023, Bao2023, Kang2023, Zheng2023, Dubois2024, WangAlign2023}. We unify these by exposing reasoning collapse on subjective industry rubrics and proposing a length-penalized mitigation together with a persona-routing blueprint.

\section{Discussion and Conclusion}
\label{sec:conclusion}

Deploying LLMs as autonomous verifiers is a critical frontier for recommender systems. Contrary to the prevailing consensus that RLVR universally enhances LLM capabilities via Chain-of-Thought, our analysis on large-scale, real-world subjective tasks reveals a starkly different reality: stripped of objective, mathematical ground truths, explicit reasoning frequently \emph{degrades} verification and suffers \textit{reasoning collapse} under standard policy optimization. To bridge the gap between deterministic mathematical reasoning and subjective, human-centric evaluation, our conditional length-penalized reward mathematically intertwines the preservation of reasoning length with strict verification accuracy---preventing context exploitation, halting collapse, and recovering (indeed surpassing) baseline performance on complex industry rubrics. This yields a scalable algorithmic patch that practitioners can adopt today.

For practitioners, these findings translate into concrete guidance. First, reasoning should be treated as a \emph{tunable}, not a default: on subjective rubrics a zero-shot verifier is often a stronger and cheaper baseline than one prompted to deliberate at length. Second, when post-training is warranted, the reward must explicitly protect the reasoning budget, since outcome-only objectives silently collapse it. Third, because the appropriate reasoning style is task- and audience-dependent, teams are better served by \emph{routing} over a small library of vetted personas than by distilling a single monolithic chain-of-thought---and any such deployment should be paired with human oversight and fairness auditing.

Solving this mechanical collapse, however, illuminated a deeper structural need: monolithic reasoning inherited from mathematical pre-training is fundamentally insufficient for subjective tasks, and a reasoning trace's efficacy is tightly coupled with its socio-linguistic framing. The path forward is therefore not merely better gradient clipping but a shift in \emph{how} reinforcement learning is applied---toward an architecture in which RLVR acts as a contextual multi-armed bandit that dynamically routes subjective queries through aligned reasoning personas rather than forcing a single reasoning mode. As consumer platforms increasingly rely on generative models for safety and evaluation at scale, we hope this work offers both a cautionary baseline and a concrete blueprint for navigating the friction of real-world subjective constraints.

\section*{Limitations}

We view this study as an encouraging first step, and several aspects naturally invite follow-up work. Our persona-driven results (Section~\ref{sec:exploratory}) are intentionally preliminary: we quantify how persona-conditioned reasoning affects verification accuracy, and we see training and deploying the full persona-routing architecture as the exciting next step rather than a solved problem. Our experiments are grounded in real, in-production rubrics and data from Netflix; this industrial grounding is a core strength, though it also means the absolute numbers reflect one deployment setting and the proprietary data cannot be publicly released. To support transferability, we report protocols, trends, and relative comparisons in detail. Our open-source post-training focuses on two widely used 7B instruction-tuned models and three representative RLVR algorithms, chosen for popularity and reproducibility; extending to larger models and additional algorithm families is straightforward future work that we expect to corroborate our findings. Because the proprietary models are accessed through APIs, their intrinsic reasoning is not directly observable and provider-side updates may shift absolute numbers over time; accordingly, our conclusions emphasize relative trends, which we expect to be more stable. Finally, we evaluate primarily with macro-F1 against human labels and leave a dedicated study of reasoning-trace faithfulness to future work. We believe these scoping choices do not affect our core findings, and each points to a concrete and promising avenue for continued research.

\section*{Ethical Considerations}

This work studies LLM verifiers for subjective, safety-relevant content moderation, which we approach with care. Our persona analysis shows that different reasoning personas yield substantially different verification outcomes. We deliberately characterize personas by reasoning disposition (e.g., communication norms and interpretive stance) rather than by protected demographic attributes, and we report only aggregate performance variation; persona conditioning should not be used to attribute viewpoints to, or stereotype, any demographic group. Because persona routing can encode or amplify bias, we recommend that any deployment be paired with fairness auditing, human oversight, and clear governance over which reasoning biases are permitted for a given task. All annotation followed the data platform's internal guidelines; the datasets contain potentially sensitive content used solely for safety and quality evaluation, and no personally identifying user data is reported. We encourage practitioners to treat automated subjective verification as decision support that augments, rather than replaces, human judgment.

\begin{acks}
During the preparation of this manuscript, the authors used a generative AI assistant for \LaTeX{} formatting and language refinement. All research contributions (including the experimental design, implementation, analysis, and conclusions) are the authors' own.
\end{acks}

\bibliographystyle{ACM-Reference-Format}
\bibliography{ref}

\appendix

\section{Detailed Experimental Setup}
\label{app:setup}
\textbf{Models.} Proprietary models are accessed through their public APIs as black boxes; for the reasoning model we use the provider's ``low'' and ``high'' reasoning-effort settings. Open-source experiments use the publicly released \texttt{Mistral-7B-Instruct} and \texttt{Qwen2.5-7B-Instruct} checkpoints.

\textbf{Evaluation protocol.} For each task we evaluate on a held-out, human-labeled test set. Single-prompt results use greedy decoding; self-consistency samples $K{=}8$ generations at temperature $0.7$ and takes the majority vote. All reported numbers are macro-F1 averaged over 5 independent runs; run-to-run standard deviation was below $0.01$ macro-F1 and did not change the qualitative trends.

\textbf{Post-training.} For each prompt we sample a group of $G{=}8$ candidate generations (rollout temperature $1.0$), score each with the binary verification reward, and optimize the group-relative advantage. We optimize with AdamW at a constant learning rate of $1\times10^{-6}$, a KL-regularization coefficient $\beta{=}0.04$, a rollout batch of $64$ prompts, a maximum generation length of $8192$ tokens, and a conditional-reward target $L_{target}{=}1000$ tokens with scaling constant $\lambda{=}2\times10^{-4}$. Each model--algorithm pair is trained for $400$ steps; we report the checkpoint with the best validation macro-F1 (averaged over 5 runs). We did not specially tune these values; a coarse search over learning rate $\in\{5\times10^{-7}, 1\times10^{-6}, 2\times10^{-6}\}$ and $\beta\in\{0.01, 0.04\}$ produced consistent observations, indicating our findings are not sensitive to the exact configuration.

\section{Task Definitions and Rubric Details}
\label{app:tasks}
Each task is governed by an internal guideline document enumerating positive and negative cases. \textbf{Task~I (Conversational Query Sensitivity)} flags queries that violate sensitivity guidelines. \textbf{Task~II (Generated Response Sensitivity)} verifies whether a generated response is sensitivity-compliant. \textbf{Task~III (Text: Response Quality)} judges whether a response text adequately and relevantly addresses the query. \textbf{Task~IV (ID: Response Quality)} judges whether the catalog title in the response addresses the query. Annotators apply the same rubric used in production, and the resulting labels serve as ground truth for both evaluation and post-training. Because the rubrics are abstract and human-centric, inter-annotator disagreement is itself informative and motivates our focus on subjective verification.

\section{Persona Synthesis Details}
\label{app:persona}
We draw $1{,}500$ personas from a public roleplay dataset~\citep{singh2026fspofewshotoptimizationsynthetic}. For each persona we construct a system prompt instructing the model to adopt that persona's reasoning style and to produce (i) a short reasoning trace and (ii) a final ``Yes''/``No'' verification, for every item in a held-out evaluation subset of a sensitivity task. We then compute macro-F1 of each persona's verifications against the human ground truth, producing the distribution in Figure~\ref{fig:persona}. To avoid stereotyping, we summarize personas only by their aggregate effect on reasoning style and verification accuracy, not by protected attributes, and report the distribution rather than identifying individual personas.

\begin{figure}[t]
    \centering
    \includegraphics[width=0.95\linewidth]{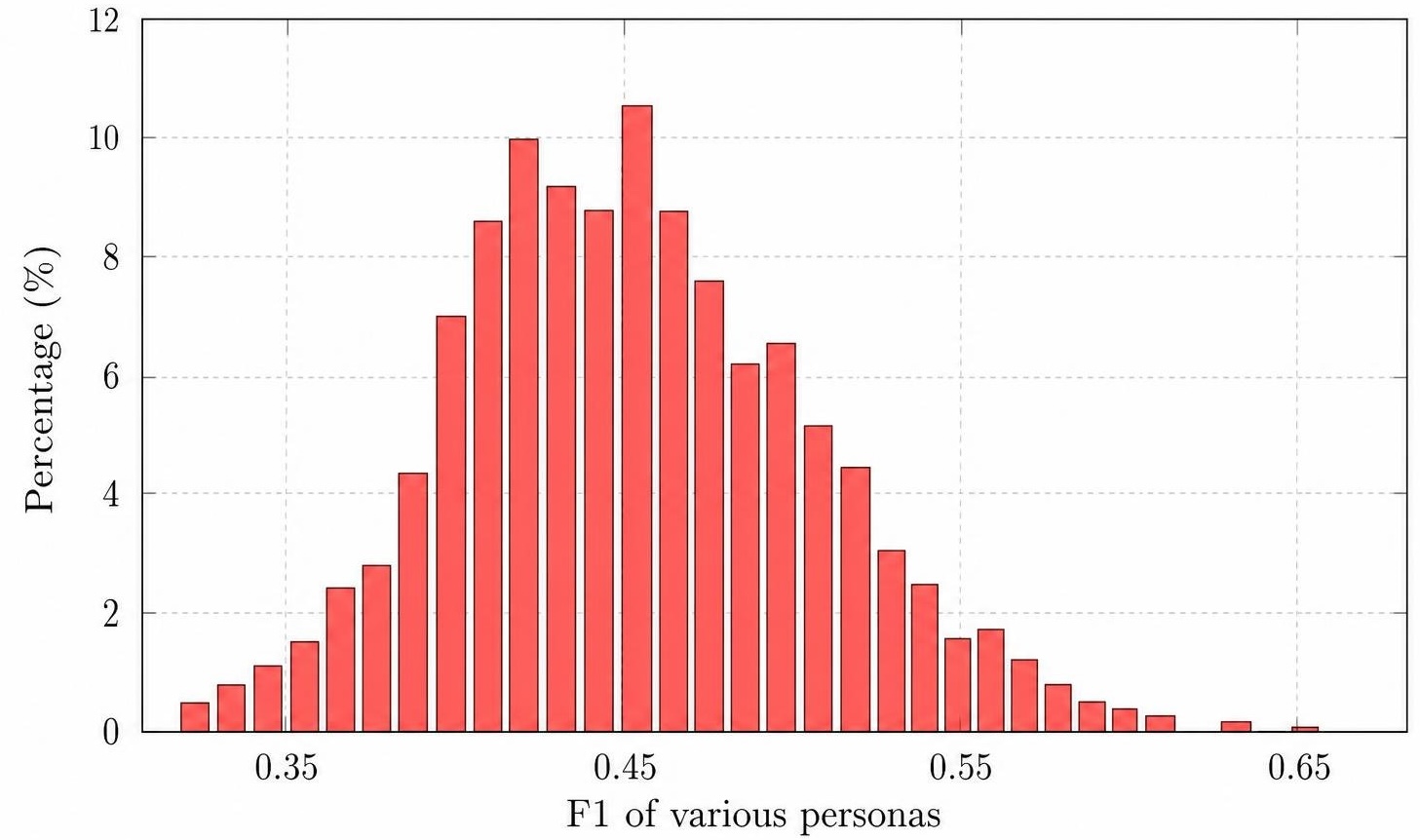}
    \caption{Persona-conditioned macro-F1 distribution on the ID: Response Quality task (1{,}500 personas). As on the sensitivity task (Figure~\ref{fig:persona}), persona choice induces a wide spread; the absolute values are lower, reflecting the harder rubric.}
    \Description{Histogram of macro-F1 across 1,500 personas on the ID: Response Quality task, a wide unimodal distribution peaked near 0.45 spanning about 0.33 to 0.65.}
    \label{fig:persona2}
\end{figure}

We repeat the analysis on a second, harder task, ID: Response Quality (Figure~\ref{fig:persona2}). Consistent with the sensitivity task in Figure~\ref{fig:persona}, persona choice again induces a wide macro-F1 spread (roughly $0.33$ to $0.65$, peaked near $0.45$); the absolute values are lower, reflecting the greater difficulty of this rubric (cf.\ Tables~\ref{tab:proprietary_performance}--\ref{tab:reward_shaping}). This corroborates that the effect of persona-conditioned reasoning is general rather than specific to one task.

\section{Extended Related Work}
\label{app:relwork}

\paragraph{Large Language Models and Reasoning.} Test-time compute via intermediate token generation, notably Chain-of-Thought (CoT), has fundamentally advanced LLM reasoning \citep{Wei2022, Kojima2022}. Extensions such as Self-Consistency \citep{Wang2022} and Tree of Thoughts \citep{Yao2023} further amplify inference-time search. Recent advancements shift toward post-training models explicitly for reasoning via reinforcement learning \citep{DeepSeekR1_2025, OpenAIo1_2024}, relying heavily on process reward models (PRMs) and outcome supervision \citep{Uesato2022, Lightman2023}. However, this literature predominantly benchmarks reasoning on objective, verifiable datasets \citep{Cobbe2021, Hendrycks2021}. Our work addresses the critical gap of how these reasoning paradigms generalize to subjective, human-aligned industry tasks.

\paragraph{Reinforcement Learning for Alignment.} Aligning LLMs with human preferences traditionally relies on Reinforcement Learning from Human Feedback (RLHF) \citep{Ziegler2019, Ouyang2022} and alternative frameworks like Constitutional AI \citep{Bai2022} and Direct Preference Optimization (DPO) \citep{Rafailov2023}. Recently, Reinforcement Learning with Verifiable Rewards (RLVR) and algorithms like GRPO \citep{Shao2024} have emerged for efficient post-training. Despite these advances, optimizing models on subjective feedback suffers from inherent limitations, including reward overoptimization and misalignment \citep{Casper2023, Gao2023}. We build upon this foundation, identifying the ``reasoning collapse'' failure mode when standard RLVR is applied to subjective rubrics, and propose a length-penalized post-training mitigation.

\paragraph{LLMs as Verifiers in Recommender Systems.} The integration of LLMs into recommender systems has scaled rapidly, transitioning from predictive modeling \citep{Bao2023, Kang2023} to broad zero-shot evaluators \citep{Hou2023}. Consequently, using ``LLM-as-a-Judge'' has become standard practice for scalable content and query verification \citep{Zheng2023, Dubois2024, WangAlign2023}. While fine-tuning generative reward models improves verification on structured tasks, applying them to subjective industry rubrics introduces high variance. To this end, this work contributes a scalable blueprint for deploying reasoning-enhanced LLM verifiers for subjective tasks without performance degradation.

\end{document}